\documentclass[10pt]{article}
\usepackage[utf8]{inputenc}
\usepackage[T1]{fontenc}
\usepackage{lmodern}
\usepackage{booktabs}
\usepackage{longtable}
\usepackage{array}
\usepackage{graphicx}
\usepackage{xcolor}
\usepackage{hyperref}
\usepackage{geometry}
\providecommand{\tightlist}{\setlength{\itemsep}{0pt}\setlength{\parskip}{0pt}}

\hypersetup{colorlinks=true,linkcolor=blue,urlcolor=blue,citecolor=blue}

\title{Small Experiments, Cheaper Decisions: A Case Study in Staged Promotion for Micro-Pretraining}
\author{{\large Felipe Chavarro Polania}\\[0.35em]{\normalsize Hewlett Packard Enterprise}\\[0.15em]{\small\texttt{felipechavarropolania@hpe.com}}}
\date{2026-06-09}

\begin{document}
\maketitle

\begin{abstract}
Short pretraining runs make many candidate recipes affordable, but they
can also over-promote configurations that only look strong at tiny
budgets. We study this tradeoff as a bounded case study in staged
promotion for a fixed single-GPU micro-pretraining runner. Here,
"micro-pretraining" means a single-node, single-GPU experimental runner
with staged wall-clock budgets, not that every run is sub-minute.
Starting from twelve candidate configurations derived from a prior
public screening study, we run budgets of \texttt{2} minutes, \texttt{5}
minutes, \texttt{10} minutes, \texttt{60} minutes, and \texttt{12} hours
on two heterogeneous host blocks: a Windows A100 path and a Linux L40S
path.

The early screens are useful but unstable: at \texttt{5} minutes, the
best Windows and Linux conditions differ, and the eventual
\texttt{12}-hour top-ranked condition is not the mean-best condition at
the replicated \texttt{10}-minute gate. Because seed ranges differ
across stages, these changes are operational promotion evidence rather
than within-seed learning-curve estimates. A replicated
\texttt{60}-minute gate then retains the bridge reference derived from
\emph{Staged Factorial Screening for Budget-Constrained
Micro-Pretraining} {[}1{]}, where it ranks first in all four host-seed
cells. In the final \texttt{12}-hour confirmation package, the bridge
reference ranks first in all four host-seed cells across seeds
\texttt{46} and \texttt{47}; the greedy comparator ranks second but does
not meet the frozen \texttt{0.010\ val\_bpb} near-equivalence rule; and
the d8/ar48 (depth-8, aspect-48) cheaper sentinel ranks third and does
not meet the frozen \texttt{0.020\ val\_bpb} mean-gap
cheaper-architecture rule.

The executed \texttt{12}-hour branch spends \texttt{144} GPU-hours, and
the full staged protocol records \texttt{169.2} training GPU-hours
including screening stages. Continuing all four \texttt{60}-minute
candidates to the same confirmation would spend \texttt{192} GPU-hours;
continuing all nine replicated \texttt{10}-minute candidates would spend
\texttt{432} GPU-hours. The latter two numbers are accounting
counterfactuals for unrun continuations, not evidence that skipped
candidates could not have overtaken the reference. The result is a
bounded cost-allocation finding, not evidence that the protocol
outperforms adaptive hyperparameter optimization or that the largest
model\textquotesingle s advantage is capacity-normalized.
\end{abstract}

\section{1. Introduction}\label{1-introduction}

Small pretraining experiments are often used as filters before heavier
training budgets are spent. The operational reason is simple: if a short
run can reject a bad recipe, the saved accelerator time can be used
elsewhere. The scientific risk is also simple: a short run can rank
configurations in an order that does not survive more time, another
seed, or another host.

In this paper, micro-pretraining means a single-node, single-GPU
experimental runner used to make small-budget pretraining decisions. It
does not mean that all stages are tiny in wall-clock time: the protocol
deliberately escalates from minute-scale screens to \texttt{12}-hour
confirmation runs.

This paper studies that tradeoff as a promotion problem. We do not ask
whether a short run can prove a configuration is globally best. We ask
whether a small, auditable, two-worker promotion schedule can keep an
observed long-horizon reference in the candidate set while identifying
plausible alternatives that do not meet frozen continuation thresholds.

The study starts from a twelve-condition matrix derived from the prior
staged factorial micro-pretraining screening campaign reported in
{[}1{]}. The bridge reference is the best reference condition carried
forward from that prior screening campaign; this paper tests whether
staged promotion retains and challenges it, not whether a naive search
discovers it from scratch. The candidates include that bridge reference,
a greedy comparator, a high-penalty control, several smaller or cheaper
variants, and local variants around the bridge region. We then run a
multi-fidelity schedule on two heterogeneous host blocks: a Windows A100
host and a Linux L40S host. Early budgets are intentionally cheap. Later
budgets are spent only after written pre-analysis gates.

Operationally, a gate observes a candidate set \texttt{S\_t}, a fixed
wall-clock budget \texttt{b\_t}, blocked host measurements, and
predeclared thresholds, then chooses a smaller set \texttt{S\_\{t+1\}}
before the next budget is spent. The object of study is not only the
final score, but whether the sequence of gates avoids over-pruning while
reducing expensive continuations.

The final result is narrower than a general optimizer claim but useful
for constrained experimentation. Early rankings are unstable enough that
a hard prune at \texttt{5} or \texttt{10} minutes would be risky.
However, carrying reference and host-sensitive candidates through a
replicated \texttt{60}-minute gate keeps the eventual \texttt{12}-hour
top-ranked condition in the promoted set. Because seed ranges change
across stages, this is retention evidence under an operational promotion
schedule, not causal evidence that budget duration alone changed the
ordering. The \texttt{12}-hour package then closes the cheaper-model
branch for this study: the bridge reference ranks first in all four
host-seed cells, while the cheaper sentinel and greedy comparator do not
meet the frozen thresholds. No \texttt{24}-hour continuation is launched
after this result.

The main contribution is therefore methodological discipline rather than
a new architecture. A staged promotion rule can reduce long-horizon
spending if it is used conservatively: broad cheap screens, replicated
intermediate gates, frozen thresholds, and explicit stopping when
plausible branches fail.

\section{2. Contributions}\label{2-contributions}

This paper makes four contributions.

\begin{enumerate}
\def\labelenumi{\arabic{enumi}.}
\tightlist
\item
  It documents a fully staged promotion protocol for this fixed
  micro-pretraining runner: smoke test, cheap screen, replicated cheap
  screen, replicated \texttt{60}-minute confirmation, and two-seed
  \texttt{12}-hour confirmation, with frozen gates and auditable budget
  accounting.
\item
  It shows that the early screen is not stable enough for aggressive
  pruning: the \texttt{5}- and \texttt{10}-minute reads are
  host-sensitive, and the eventual \texttt{12}-hour top-ranked condition
  is not the mean-best condition at \texttt{10} minutes. Because later
  stages use different seed ranges, this is operational promotion
  evidence rather than a within-seed duration effect.
\item
  It shows that a conservative promotion rule can still retain the
  long-horizon reference derived from \emph{Staged Factorial Screening
  for Budget-Constrained Micro-Pretraining} {[}1{]} in the promoted set:
  the bridge reference passes every gate and ranks first in all four
  \texttt{12}-hour host-seed cells under this fixed wall-clock protocol,
  a comparison that is capacity-confounded because the bridge reference
  is also the largest final condition.
\item
  It provides budget accounting for the stopping decision: the executed
  \texttt{12}-hour confirmation uses \texttt{144} GPU-hours, compared
  with \texttt{192} GPU-hours for continuing all four \texttt{60}-minute
  candidates and \texttt{432} GPU-hours for continuing all nine
  \texttt{10}-minute candidates. These comparison budgets are accounting
  counterfactuals, not observed outcomes for skipped continuations.
\end{enumerate}

\section{3. Related Work}\label{3-related-work}

Hyperparameter optimization under finite budgets is the closest
methodological context. Hyperband formulates hyperparameter optimization
as adaptive resource allocation with early stopping over randomly
sampled configurations {[}2{]}. ASHA extends successive-halving-style
promotion to massively parallel settings {[}3{]}. BOHB combines
model-based search with bandit-style budget allocation {[}4{]}. These
methods motivate the idea that not every configuration should receive
the largest budget. This paper does not claim to outperform those
methods. Instead, it studies a small, manually auditable promotion
protocol for a fixed runner, two workers, and a narrow candidate matrix.

The distinction is operational. Hyperband, ASHA, BOHB, and Bayesian
optimization are automated search procedures for broader optimization
problems. The protocol studied here is a practitioner-in-the-loop
decision record: it freezes gates, preserves reference and control
value, and explains why particular continuations are stopped. It is
complementary to automated HPO rather than a replacement for it.

Reporting practices are also central. Dodge et al. argue that final test
scores alone are insufficient and recommend showing validation
performance as a function of compute budget {[}5{]}. Our figures follow
that principle: the paper reports stage trajectories, host-seed cells,
and budget counterfactuals rather than only the final \texttt{12}-hour
winner.

Small-scale pretraining decisions are increasingly important because
full pretraining comparisons are expensive. DataDecide studies how well
small experiments predict larger pretraining choices across many corpora
and scales {[}6{]}. Optimizer-comparison work has also emphasized that
rankings can flip with training scale, tuning effort, and evaluation
timing {[}7{]}. Those warnings are directly relevant here: we treat
early screens as candidate-generation mechanisms, not as proof of
long-horizon quality.

\section{4. Experimental Setup}\label{4-experimental-setup}

\subsection{4.1 Runner And Hosts}\label{41-runner-and-hosts}

All experiments use a fixed micro-pretraining runner derived from the
prior screening branch reported in {[}1{]} and instrumented for the
staged-promotion study. The runner reports final validation bits per
byte (\texttt{val\_bpb}, lower is better), parameter count, peak VRAM,
total tokens, training seconds, and final checkpoint path. We use
\texttt{val\_bpb} rather than perplexity because it is a direct
compression-style validation loss for the fixed byte/token stream and
remains comparable across wall-clock-limited runs where models process
different token counts.

\texttt{val\_bpb\ =\ -\ (1\ /\ (N\ log\ 2))\ *\ sum\_\{i=1\}\^{}\{N\}\ log\ p(x\_i\ \textbar{}\ x\_\{\textless{}i\})},

where \texttt{log} denotes the natural logarithm, \texttt{N} is the
number of evaluated validation bytes/tokens under the fixed tokenizer
stream, and lower values indicate better compression of the validation
stream. The primary experiment path uses two heterogeneous host blocks:

\begin{longtable}[]{@{}lll@{}}
\toprule
host block & operating path & accelerator \\
\midrule
\endhead
\bottomrule
\endlastfoot
Windows & Windows runner path & NVIDIA A100 40GB \\
Linux & Linux runner path & NVIDIA L40S \\
\end{longtable}

The hosts are not treated as identical replicas. They are blocked
observations used to test whether the same promotion decision remains
directionally visible after changing operating path and accelerator. We
use heterogeneous hardware to test whether promotion decisions survive
the most conservative host change available in this environment, while
acknowledging that the A100-L40S difference conflates architecture,
operating system, driver stack, and filesystem path. The descriptive
standard deviations reported in Section 5.4 should be read against this
composite-block caveat.

The frozen runner uses locally cached training shards from
\texttt{karpathy/climbmix-400b-shuffle}, with
\texttt{shard\_06542.parquet} pinned as the validation shard.
Tokenization uses a rustbpe-trained, tiktoken-compatible BPE with
vocabulary \texttt{8192}. All runs use context length \texttt{2048} and
report final \texttt{val\_bpb} over \texttt{40\ *\ 524288} validation
tokens from the pinned shard. Appendix A records the dataset URL, shard
identifier, tokenizer artifacts, source snapshot, and reproducibility
bundle contents.

The runner fixes the run seed with Python and PyTorch seed calls and
records the seed in each summary. It does not claim bitwise
deterministic replay across GPU architectures: deterministic PyTorch
algorithms are not enabled,
\texttt{torch.set\_float32\_matmul\_precision("high")} is used, and
CUDA/cuDNN kernels may differ between the A100 and L40S paths. The seed
design therefore supports repeated operational reads, not exact binary
replay.

\subsection{4.2 Candidate Matrix}\label{42-candidate-matrix}

The starting matrix contains twelve conditions. The prior \emph{Staged
Factorial Screening for Budget-Constrained Micro-Pretraining} study
{[}1{]} screened depth, aspect-ratio, and learning-rate settings and
produced the bridge reference, a greedy comparator, and a high-penalty
control. The remaining conditions fill local variants and smaller-model
cells around that region to test whether staged promotion retains or
rejects plausible alternatives. The roles are predeclared so that
promotion can retain reference and control value instead of only
selecting the current best short-budget row. The short labels are used
in later compact result tables. Exact condition identifiers are internal
reproducibility IDs and are carried in the ancillary matrix.

\begin{longtable}[]{@{}llrrrrr@{}}
\toprule
label & condition id & role & depth & aspect & matrix lr & batch \\
\midrule
\endhead
\bottomrule
\endlastfoot
bridge & \texttt{p06\_bridge\_best} & reference best & 8 & 64 & 0.05 &
262144 \\
greedy & \texttt{p06\_greedy\_winner} & search comparator & 6 & 72 &
0.03 & 262144 \\
control & \texttt{p06\_control} & high-penalty control & 8 & 48 & 0.03 &
524288 \\
c03 & \texttt{p06\_best\_c03} & small reference & 6 & 48 & 0.05 &
262144 \\
c01 & \texttt{p06\_best\_c01} & small reference & 6 & 48 & 0.03 &
262144 \\
bridge-d6 & \texttt{p06\_bridge\_d6\_ar64} & shallow bridge & 6 & 64 &
0.05 & 262144 \\
d4/ar48 & \texttt{p06\_small\_d4\_ar48\_lr05} & aggressive small & 4 &
48 & 0.05 & 262144 \\
d4/ar64 & \texttt{p06\_small\_d4\_ar64\_lr05} & centered small & 4 & 64
& 0.05 & 262144 \\
d4/ar72 & \texttt{p06\_small\_d4\_ar72\_lr03} & shallow wide & 4 & 72 &
0.03 & 262144 \\
d6/ar64 & \texttt{p06\_d6\_ar64\_lr03} & local variant & 6 & 64 & 0.03 &
262144 \\
d8/ar48 & \texttt{p06\_d8\_ar48\_lr05} & local variant & 8 & 48 & 0.05 &
262144 \\
d4/highbatch & \texttt{p06\_small\_highbatch\_d4\_ar64} & cheap
high-batch control & 4 & 64 & 0.05 & 524288 \\
\end{longtable}

\subsection{4.3 Promotion Schedule}\label{43-promotion-schedule}

The experiment uses staged wall-clock budgets. Each gate is written
before the next expensive stage.

\begin{longtable}[]{@{}lrrrrrl@{}}
\toprule
stage & candidates & seeds & hosts & budget per condition & budgeted
GPU-hours & purpose \\
\midrule
\endhead
\bottomrule
\endlastfoot
Stage 0 & 3 & 1 & 2 & \texttt{2} min & \texttt{0.2} & instrumentation
smoke test \\
Stage 1A & 12 & 1 & 2 & \texttt{5} min & \texttt{2.0} & cheap early
screen \\
Stage 1B & 12 & 1 & 2 & \texttt{10} min & \texttt{4.0} & first longer
cheap screen \\
Stage 1C & 9 & 1 & 2 & \texttt{10} min & \texttt{3.0} & seed-43 top-9
replication \\
Stage 2A & 4 & 1 & 2 & \texttt{60} min & \texttt{8.0} & seed-44
confirmation \\
Stage 2B & 4 & 1 & 2 & \texttt{60} min & \texttt{8.0} & seed-45
confirmation \\
Stage 3 & 3 & 1 & 2 & \texttt{12} h & \texttt{72.0} & seed-46
long-horizon test \\
Stage 3B & 3 & 1 & 2 & \texttt{12} h & \texttt{72.0} & seed-47
confirmation \\
\end{longtable}

The final \texttt{12}-hour branch therefore spends \texttt{144}
GPU-hours. The observed recorded training time across all remote result
summaries is \texttt{169.2} GPU-hours, including earlier screens and
confirmation stages. The unrounded internal accounting value is
\texttt{169.214}, computed from per-run training seconds. These are
training-time accounting numbers; they do not include queueing, launch
overhead, or human supervision time.

\subsection{4.4 Frozen Decision Rules}\label{44-frozen-decision-rules}

The first cheap screens were used conservatively. After the \texttt{5}-
and \texttt{10}-minute seed-42 screens, host rank agreement was still
low, so the next action was not a \texttt{60}-minute jump. Instead, we
repeated the top-9 subset at \texttt{10} minutes with seed \texttt{43}.

After the replicated \texttt{10}-minute screen, four conditions were
promoted to \texttt{60} minutes:

\begin{longtable}[]{@{}ll@{}}
\toprule
condition & reason \\
\midrule
\endhead
\bottomrule
\endlastfoot
d8/ar48 & best robust absolute performer at replicated \texttt{10}
minutes \\
d6/ar64 & cheap sentinel within the short-budget tolerance \\
bridge & predeclared bridge reference \\
greedy & predeclared greedy comparator \\
\end{longtable}

After the replicated \texttt{60}-minute gate, three conditions were
promoted to \texttt{12} hours:

\begin{longtable}[]{@{}ll@{}}
\toprule
condition & reason \\
\midrule
\endhead
\bottomrule
\endlastfoot
bridge & ranked first in every \texttt{60}-minute host-seed cell \\
greedy & greedy comparator and near-best on Windows \\
d8/ar48 & best cheaper-architecture sentinel \\
\end{longtable}

The frozen Stage 3B rule was:

\begin{longtable}[]{@{}ll@{}}
\toprule
condition & pass criterion \\
\midrule
\endhead
\bottomrule
\endlastfoot
bridge & remains best on both hosts and both \texttt{12}-hour seeds \\
greedy & within \texttt{0.010\ val\_bpb} of bridge in all
\texttt{12}-hour host-seed cells \\
d8/ar48 & within \texttt{0.020\ val\_bpb} of bridge on mean
\texttt{12}-hour gap \\
\end{longtable}

The aggregation asymmetry is intentional but should be read as a policy
choice, not a statistical discovery. The greedy comparator is a
same-class comparator, so the rule required near-equivalence in every
host-seed cell. The d8/ar48 condition is a cheaper-architecture
sentinel, so the rule allowed a wider mean-gap tolerance to ask whether
a smaller branch was "good enough" on average. The \texttt{0.010} and
\texttt{0.020\ val\_bpb} thresholds are predeclared policy bands for
this fixed runner and were used as stopping rules, not as general
significance thresholds; the sensitivity table lets readers assess how
robust the decisions are to nearby alternatives.

The public ancillary bundle includes both the written pre-analysis
records and a threshold-sensitivity table. At \texttt{12} hours, greedy
would pass under a looser \texttt{0.020} mean-gap convention but fails
the declared all-cell \texttt{0.010} max-gap rule; d8/ar48 fails both
mean-gap and max-gap checks through \texttt{0.030}. At \texttt{60}
minutes, both greedy and d8/ar48 pass a \texttt{0.020} mean-gap screen
but fail a corresponding max-gap check, consistent with their retention
for \texttt{12}-hour confirmation under predeclared roles rather than
acceptance as final. The paper therefore treats the thresholds as frozen
stopping rules rather than general superiority tests.

\subsection{4.5 Analysis Policy}\label{45-analysis-policy}

The primary endpoint is final \texttt{val\_bpb} at the end of the
allocated wall-clock budget. Because the long-horizon read has only two
seeds per final condition and two heterogeneous host blocks, we do not
report p-values for the \texttt{12}-hour comparison. We report ranks,
gaps from the bridge reference, token counts, parameter counts, peak
VRAM, and budget accounting.

Hosts are treated as blocks. Seeds are treated as repeated
initialization/data-order settings within the fixed runner. The paper
uses descriptive thresholds because those thresholds were the promotion
criteria, not because the final confirmation package has enough
independent observations for broad inference.

\section{5. Results}\label{5-results}

\subsection{5.1 Stage 0: Instrumentation Smoke
Test}\label{51-stage-0-instrumentation-smoke-test}

Stage 0 verified that the instrumented runner produced parseable
summaries and final metrics on both hosts.

\begin{longtable}[]{@{}lrrrrr@{}}
\toprule
condition & mean \texttt{val\_bpb} & Windows & Linux & params M & tokens
M \\
\midrule
\endhead
\bottomrule
\endlastfoot
d4/ar64 & \texttt{1.274661} & \texttt{1.265988} & \texttt{1.283334} &
\texttt{11.534472} & \texttt{82.444288} \\
greedy & \texttt{1.467645} & \texttt{1.377154} & \texttt{1.558135} &
\texttt{39.846284} & \texttt{28.835840} \\
bridge & \texttt{1.569651} & \texttt{1.478259} & \texttt{1.661044} &
\texttt{50.332176} & \texttt{20.316160} \\
\end{longtable}

This stage is not used for scientific ranking or screening. It is a pure
instrumentation check. It shows why fixed-time comparisons need token
accounting: smaller models can process many more tokens inside the same
wall-clock budget.

\subsection{5.2 Stage 1: Cheap Screens Are Useful But
Unstable}\label{52-stage-1-cheap-screens-are-useful-but-unstable}

At \texttt{5} minutes, the best Windows and Linux conditions differ. The
Windows winner is bridge-d6; the Linux winner is d8/ar48. The cross-host
rank agreement is low, so an aggressive prune at this point would be
risky.

\begin{longtable}[]{@{}lrrr@{}}
\toprule
condition & Windows rank & Linux rank & mean \texttt{val\_bpb} \\
\midrule
\endhead
\bottomrule
\endlastfoot
bridge-d6 & 1 & 4 & \texttt{1.185989} \\
d8/ar48 & 5 & 1 & \texttt{1.180862} \\
c03 & 2 & 7 & \texttt{1.191919} \\
d4/ar48 & 8 & 2 & \texttt{1.190858} \\
d6/ar64 & 4 & 6 & \texttt{1.192230} \\
d4/ar64 & 9 & 3 & \texttt{1.191005} \\
c01 & 3 & 9 & \texttt{1.198454} \\
d4/ar72 & 6 & 10 & \texttt{1.203480} \\
d4/highbatch & 11 & 5 & \texttt{1.216565} \\
bridge & 10 & 8 & \texttt{1.211893} \\
greedy & 7 & 11 & \texttt{1.214838} \\
control & 12 & 12 & \texttt{1.264903} \\
\end{longtable}

The replicated \texttt{10}-minute top-9 screen gives a more useful but
still host-sensitive read:

\begin{longtable}[]{@{}lrrrr@{}}
\toprule
condition & mean \texttt{val\_bpb} & mean rank & best-worst rank & gap
from best \\
\midrule
\endhead
\bottomrule
\endlastfoot
d8/ar48 & \texttt{1.110701} & \texttt{3.25} & 1-6 & \texttt{0.000000} \\
d6/ar64 & \texttt{1.122445} & \texttt{3.25} & 3-4 & \texttt{0.011744} \\
bridge-d6 & \texttt{1.122590} & \texttt{3.25} & 2-5 &
\texttt{0.011889} \\
bridge & \texttt{1.122643} & \texttt{4.50} & 2-7 & \texttt{0.011942} \\
c01 & \texttt{1.126101} & \texttt{5.00} & 3-7 & \texttt{0.015400} \\
greedy & \texttt{1.129024} & \texttt{5.00} & 1-9 & \texttt{0.018323} \\
c03 & \texttt{1.127013} & \texttt{6.00} & 5-7 & \texttt{0.016313} \\
control & \texttt{1.143036} & \texttt{6.25} & 4-8 & \texttt{0.032335} \\
d4/ar48 & \texttt{1.152692} & \texttt{8.50} & 8-9 & \texttt{0.041991} \\
\end{longtable}

The final \texttt{12}-hour top-ranked condition, the bridge reference,
is not the mean-best condition at this gate. The mean-best condition at
this gate is d8/ar48 (\texttt{33.0M} parameters), which later ranks
third in the final \texttt{12}-hour confirmation package. It remains in
the promoted set because the promotion rule includes predeclared
reference/control value rather than only short-budget rank. Because
subsequent stages use new seeds, this observation should be read as a
conservative promotion warning, not as a within-seed learning-curve
reversal.

Figure 1 summarizes the staged funnel. It shows the key design choice:
spend cheap budgets broadly, then narrow before the expensive
\texttt{12}-hour branch.

\begin{figure}
\centering
\includegraphics[width=\linewidth,height=0.80\textheight,keepaspectratio]{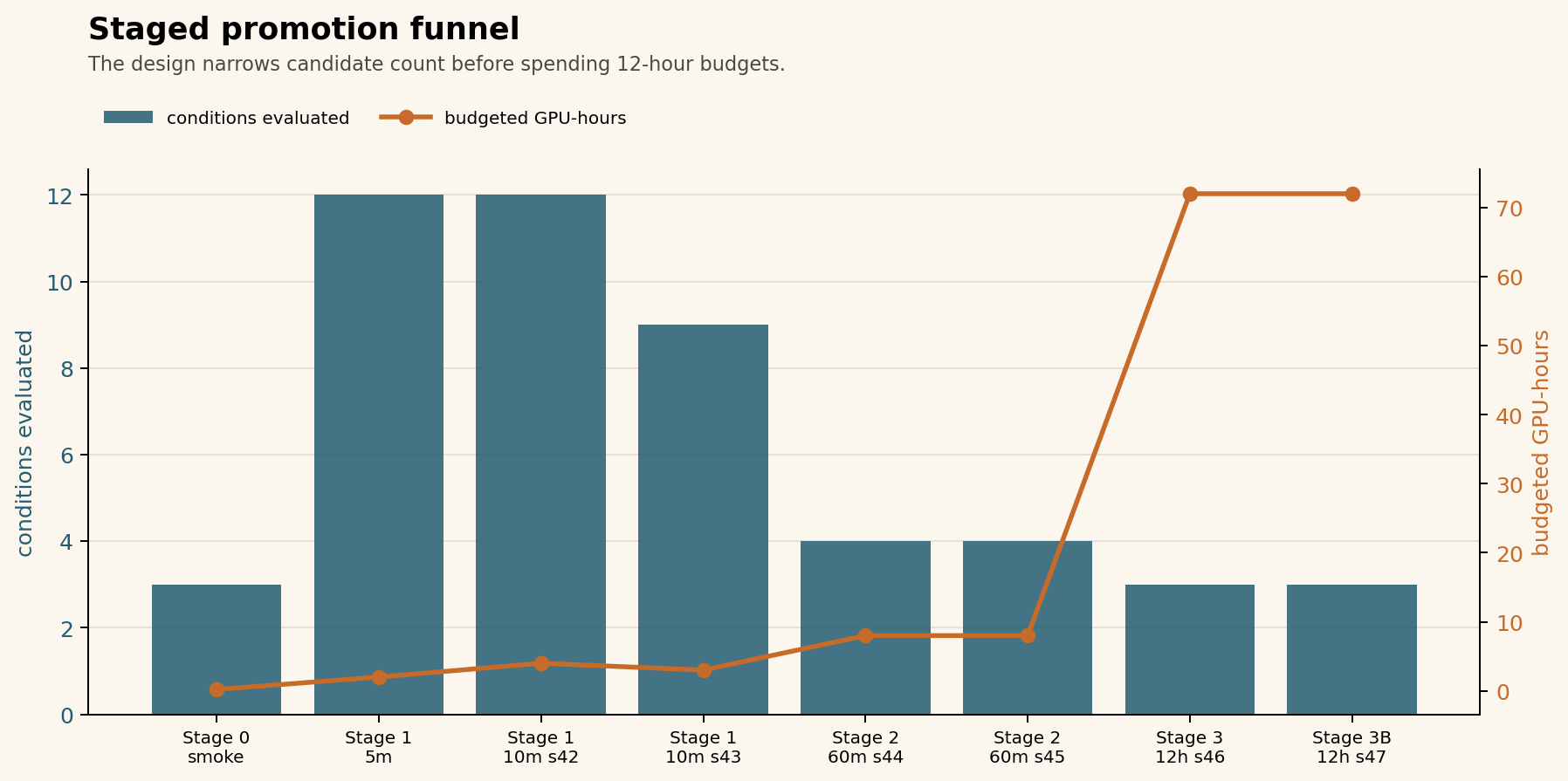}
\caption{Staged promotion funnel. The design narrows candidate count
before spending 12-hour budgets.}
\end{figure}

\subsection{5.3 Stage 2: Replicated 60-Minute Gate Retains The
Bridge}\label{53-stage-2-replicated-60-minute-gate-retains-the-bridge}

The replicated \texttt{60}-minute gate gives a stronger promotion read
under seeds \texttt{44} and \texttt{45}. The bridge reference ranks
first in all four host-seed cells. Because these are not the same seeds
used at \texttt{10} minutes, the result should be read as stage-gate
evidence for retention, not as proof that duration alone changed the
ordering.

\begin{longtable}[]{@{}lrrrrrr@{}}
\toprule
condition & mean \texttt{val\_bpb} & sd \texttt{val\_bpb} & mean gap &
max gap & params M & tokens M \\
\midrule
\endhead
\bottomrule
\endlastfoot
bridge & \texttt{0.987940} & \texttt{0.017719} & \texttt{0.000000} &
\texttt{0.000000} & \texttt{50.332176} & \texttt{609.812480} \\
greedy & \texttt{1.004475} & \texttt{0.029165} & \texttt{0.016535} &
\texttt{0.026927} & \texttt{39.846284} & \texttt{790.626304} \\
d8/ar48 & \texttt{1.007118} & \texttt{0.010141} & \texttt{0.019178} &
\texttt{0.026238} & \texttt{33.030544} & \texttt{850.001920} \\
d6/ar64 & \texttt{1.025509} & \texttt{0.018875} & \texttt{0.037569} &
\texttt{0.039608} & \texttt{26.345772} & \texttt{1081.081856} \\
\end{longtable}

This gate drops the weakest of the four promoted conditions and sends
three final conditions to \texttt{12} hours: bridge, greedy comparator,
and the best cheaper-architecture sentinel.

Figure 2 shows the gap trajectory across the operational gate sequence.
The short-budget leader does not remain competitive at \texttt{12}
hours. The bridge reference separates in the later gate reads, with the
seed-budget confound noted above.

\begin{figure}
\centering
\includegraphics[width=\linewidth,height=0.80\textheight,keepaspectratio]{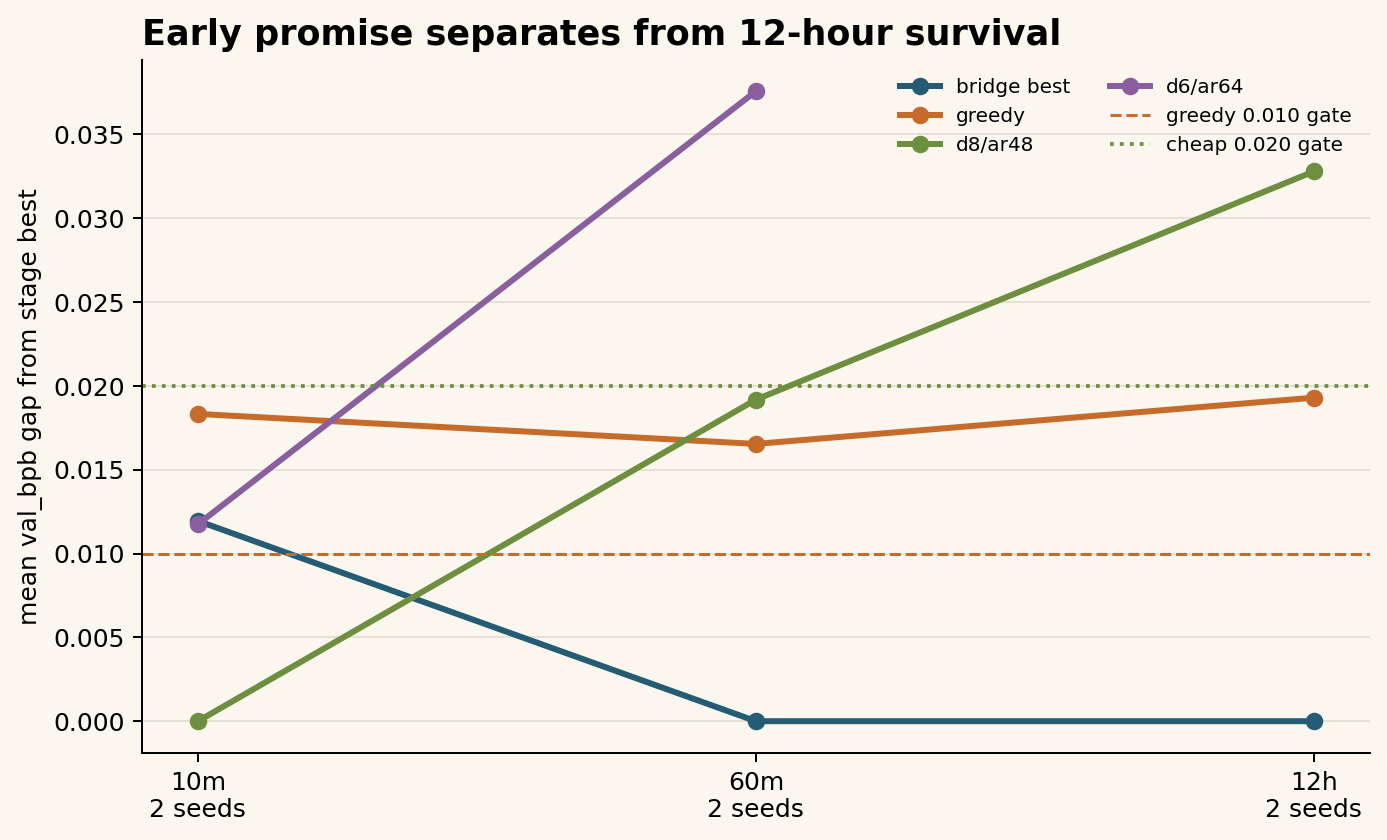}
\caption{Early promise separates from 12-hour survival. Lines show mean
\texttt{val\_bpb} gap from the stage best across the replicated
\texttt{10}-minute, replicated \texttt{60}-minute, and two-seed
\texttt{12}-hour reads; stages use different seed ranges, so the
trajectory is operational promotion evidence rather than a within-seed
learning curve.}
\end{figure}

\subsection{5.4 Stage 3: 12-Hour Confirmation Retains The Bridge
Reference}\label{54-stage-3-12-hour-confirmation-retains-the-bridge-reference}

The final confirmation package runs three conditions at \texttt{12}
hours on both hosts with seeds \texttt{46} and \texttt{47}. The bridge
ranks first in all four host-seed cells.

\begin{longtable}[]{@{}lrrrrrrr@{}}
\toprule
condition & first-rank cells & mean \texttt{val\_bpb} & sd
\texttt{val\_bpb} & mean gap & max gap & params M & tokens M \\
\midrule
\endhead
\bottomrule
\endlastfoot
bridge & 4/4 & \texttt{0.931915} & \texttt{0.006400} & \texttt{0.000000}
& \texttt{0.000000} & \texttt{50.332176} & \texttt{7275.347968} \\
greedy & 0/4 & \texttt{0.951208} & \texttt{0.010326} & \texttt{0.019294}
& \texttt{0.022927} & \texttt{39.846284} & \texttt{9428.336640} \\
d8/ar48 & 0/4 & \texttt{0.964701} & \texttt{0.004437} &
\texttt{0.032786} & \texttt{0.036890} & \texttt{33.030544} &
\texttt{10090.119168} \\
\end{longtable}

The standard deviations are descriptive across the four host-seed cells.
Given the composite host block, this variation combines seed effects
with host-block effects and is not a seed-only dispersion estimate.

The frozen thresholds are not met by either alternative. The greedy
comparator fails the \texttt{0.010\ val\_bpb} all-cell near-equivalence
rule because its maximum gap is \texttt{0.022927} and its mean gap is
\texttt{0.019294}. The d8/ar48 cheaper sentinel fails the
cheaper-architecture rule because its mean gap is \texttt{0.032786},
above the \texttt{0.020} threshold.

The threshold sensitivity is:

\begin{longtable}[]{@{}lrlll@{}}
\toprule
check & observed gap & declared rule & decision & sensitivity note \\
\midrule
\endhead
\bottomrule
\endlastfoot
greedy max gap & \texttt{0.022927} & max gap
\texttt{\textless{}=\ 0.010} & fail & would pass only at a looser
\texttt{0.030} max-gap rule \\
greedy mean gap & \texttt{0.019294} & diagnostic only & not primary &
would pass a looser \texttt{0.020} mean-gap screen \\
d8/ar48 mean gap & \texttt{0.032786} & mean gap
\texttt{\textless{}=\ 0.020} & fail & fails through \texttt{0.030} \\
d8/ar48 max gap & \texttt{0.036890} & diagnostic only & fail & fails
through \texttt{0.030} \\
\end{longtable}

This result is also capacity-confounded. The bridge reference has
\texttt{50.3M} parameters, compared with \texttt{39.8M} for the greedy
comparator and \texttt{33.0M} for d8/ar48. In the \texttt{12}-hour final
set, parameter count and final \texttt{val\_bpb} move in the same
direction while the smaller models process more tokens, so the result
should not be read as a capacity-normalized recipe comparison. It shows
that the promoted bridge reference remained the strongest observed
condition under this wall-clock protocol, not that the promotion rule
separated workflow quality from model size.

The paired host-seed values are:

\begin{longtable}[]{@{}lrrrr@{}}
\toprule
host & seed & bridge & greedy & d8/ar48 \\
\midrule
\endhead
\bottomrule
\endlastfoot
Windows & 46 & \texttt{0.926470} & \texttt{0.941986} &
\texttt{0.959303} \\
Linux & 46 & \texttt{0.937029} & \texttt{0.959956} &
\texttt{0.966803} \\
Windows & 47 & \texttt{0.926290} & \texttt{0.942552} &
\texttt{0.963180} \\
Linux & 47 & \texttt{0.937869} & \texttt{0.960339} &
\texttt{0.969518} \\
\end{longtable}

Figure 3 shows the same result as a host-seed confirmation plot.

\begin{figure}
\centering
\includegraphics[width=\linewidth,height=0.80\textheight,keepaspectratio]{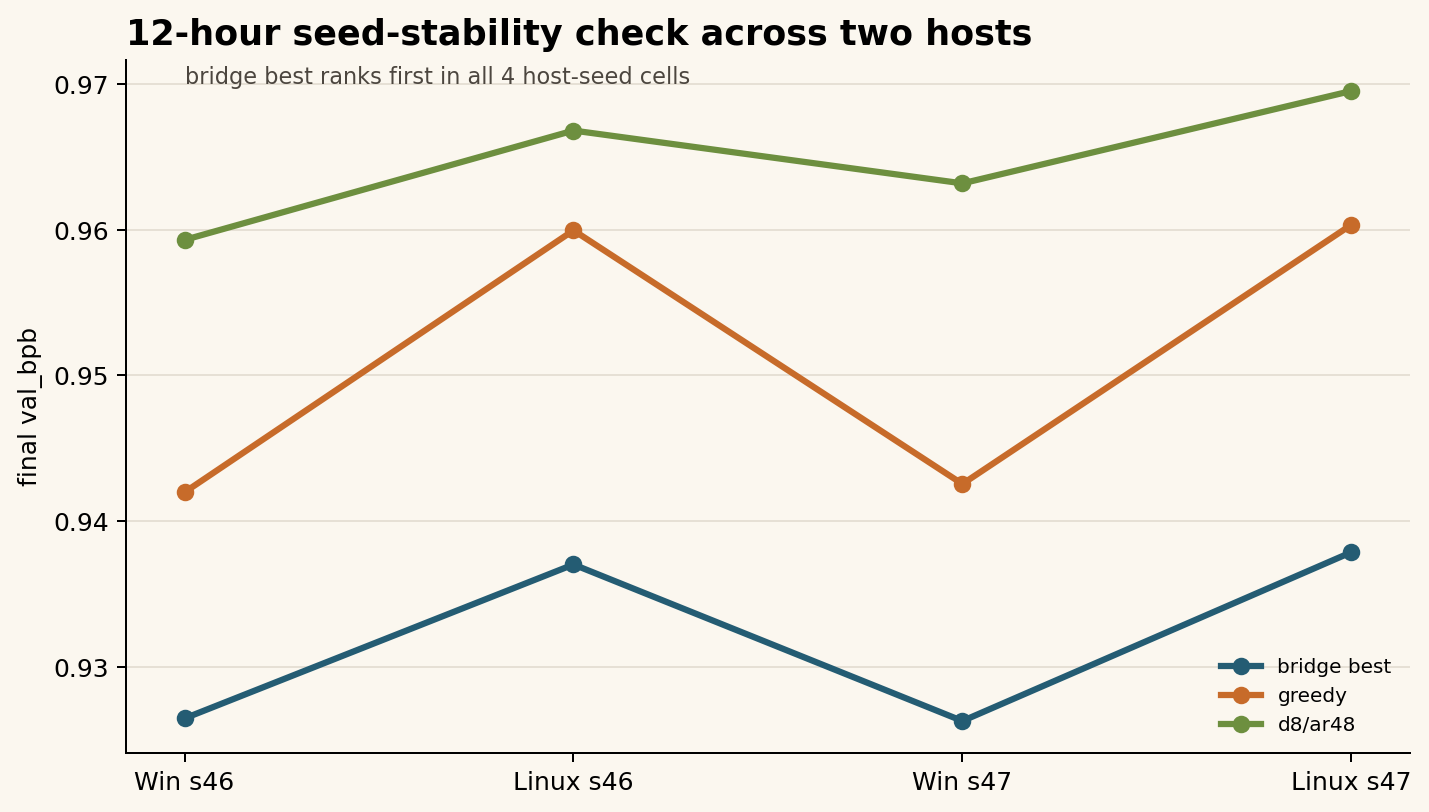}
\caption{12-hour seed-stability check across two hosts. The bridge
condition ranks first in all four host-seed cells.}
\end{figure}

\clearpage
\subsection{5.5 Cost-Quality Frontier}\label{55-cost-quality-frontier}

The final result is not a cheap-model victory. The smaller candidates
process more tokens within the same wall-clock budget, but they do not
match the bridge in final \texttt{val\_bpb}.

Figure 4 makes that tradeoff explicit. Bubble area tracks tokens
processed. The \texttt{33.0M}-parameter d8/ar48 candidate processes
about \texttt{10.09B} tokens across the four host-seed cells, compared
with about \texttt{7.28B} for the \texttt{50.3M}-parameter bridge, but
its final mean \texttt{val\_bpb} is worse by \texttt{0.032786}.

\begin{figure}
\centering
\includegraphics[width=\linewidth,height=0.80\textheight,keepaspectratio]{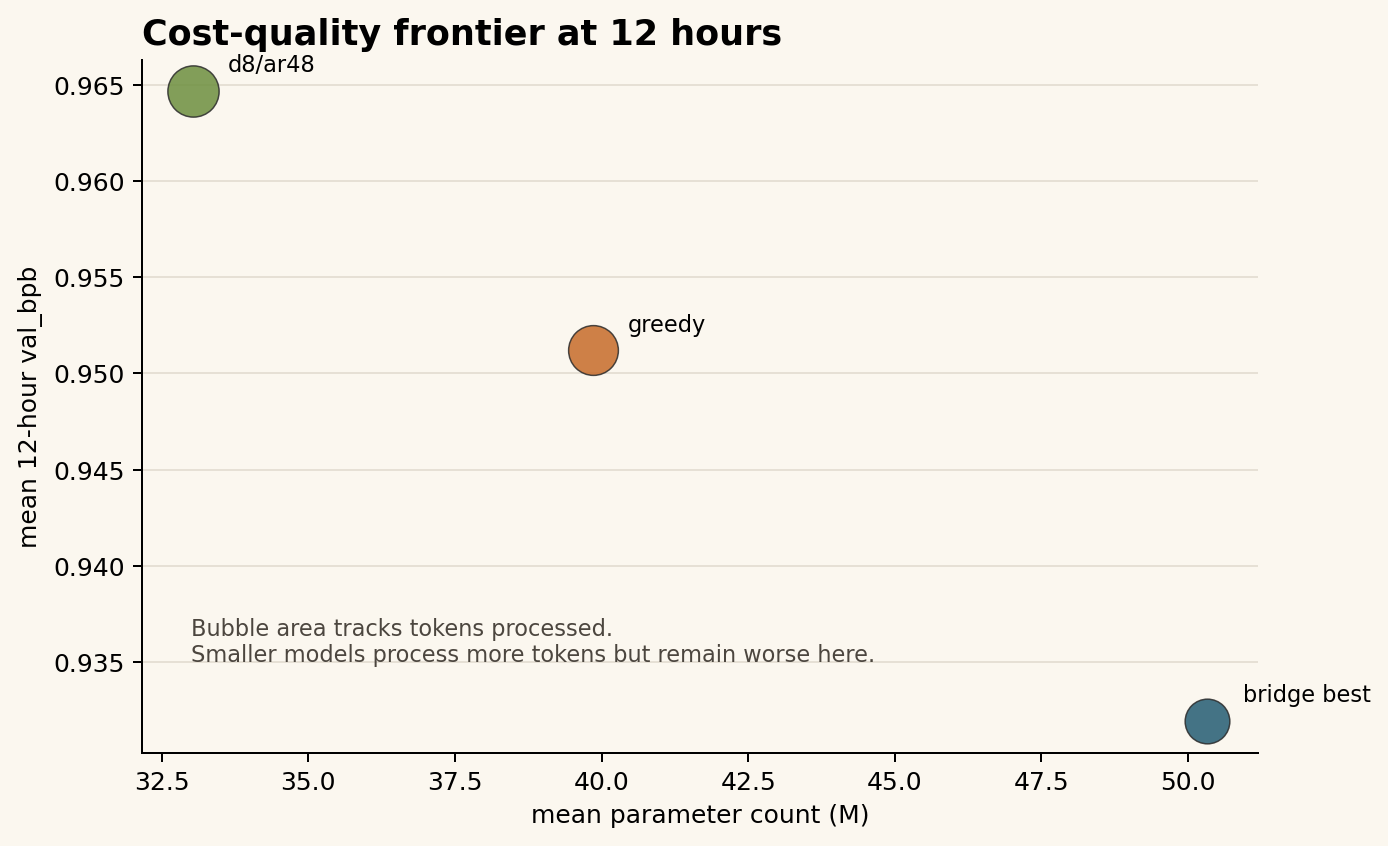}
\caption{Cost-quality frontier at 12 hours. Smaller models process more
tokens but remain worse in final \texttt{val\_bpb} in this final set.}
\end{figure}

\clearpage
\subsection{5.6 Budget Counterfactual}\label{56-budget-counterfactual}

The primary cost result is the stopping decision. The final branch
spends \texttt{144} GPU-hours on three conditions across two seeds and
two hosts. Continuing all four \texttt{60}-minute candidates would spend
\texttt{192} GPU-hours. Continuing all nine replicated
\texttt{10}-minute candidates would spend \texttt{432} GPU-hours.

\begin{longtable}[]{@{}lrr@{}}
\toprule
scenario & \texttt{12}-hour confirmation GPU-hours & added GPU-hours vs
executed branch \\
\midrule
\endhead
\bottomrule
\endlastfoot
executed three-condition branch & \texttt{144} & \texttt{0} \\
continue all four \texttt{60}-minute candidates & \texttt{192} &
\texttt{48} \\
continue all nine \texttt{10}-minute candidates & \texttt{432} &
\texttt{288} \\
\end{longtable}

These numbers are accounting counterfactuals, not observed outcomes for
the skipped continuations. They show how much long-horizon budget the
promotion rule avoided spending after plausible branches did not meet
the frozen thresholds. They should not be read as evidence that the
skipped continuations could not have won or failed to improve.

\begin{figure}
\centering
\includegraphics[width=\linewidth,height=0.80\textheight,keepaspectratio]{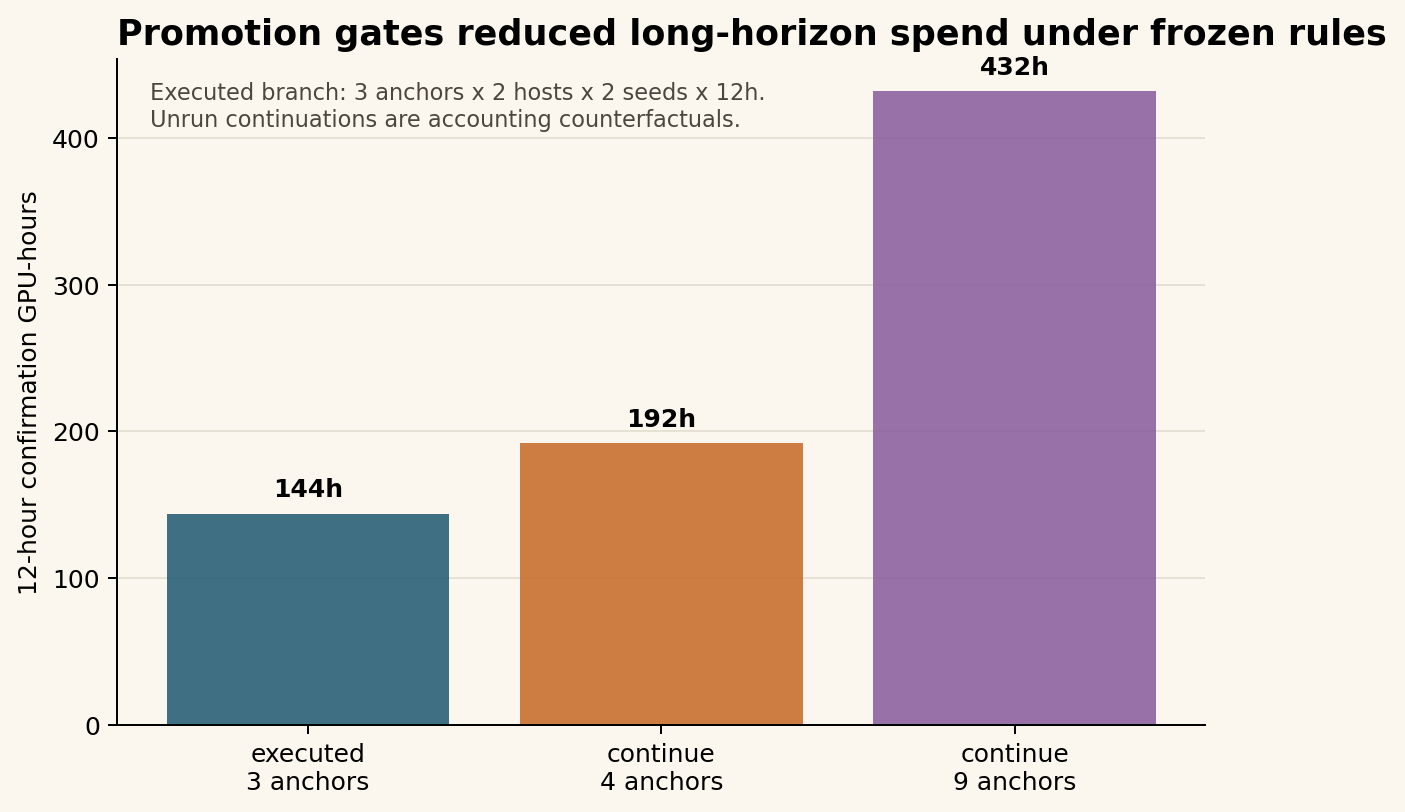}
\caption{Promotion gates reduced long-horizon spend under frozen
stopping rules. The executed branch uses 144 GPU-hours; continuing all
top-9 candidates would spend 432 GPU-hours as an accounting
counterfactual.}
\end{figure}

\section{6. Discussion}\label{6-discussion}

The result supports a conservative use of short experiments. The early
screens are valuable because they expose promising regions and obvious
failures, but they are not stable enough for final claims. The bridge
condition is not the mean-best condition at the replicated
\texttt{10}-minute gate. If the procedure had selected only the
short-budget winner, the final \texttt{12}-hour reference would have
been at risk of being dropped.

Within this seed-confounded staged design, the \texttt{60}-minute gate
is the earliest stage where the bridge reference ranks first in all
host-seed cells under the seeds used at that stage. It is still much
cheaper than a \texttt{12}-hour continuation, and it gives a better
basis for spending the later budget on a smaller anchor set, while not
proving that duration alone caused the ordering change.

The failed cheaper branch is scientifically useful. The d8/ar48
candidate is the clearest example of why cheap screens should be treated
as candidate-generation mechanisms rather than final evidence. It is the
best absolute performer at the replicated \texttt{10}-minute gate,
remains close enough at \texttt{60} minutes to justify a
\texttt{12}-hour check, and then ranks third in the final confirmation
package. Its narrow aspect ratio and smaller parameter count allow more
tokens under the fixed wall-clock budget, but that throughput advantage
does not translate into lower final \texttt{val\_bpb} in this final set.
The result does not show that smaller models cannot win. It shows that
this particular cheaper branch did not meet the frozen "good enough"
criterion, so the paper should stop instead of extending it to
\texttt{24} hours.

This is also why the paper should not be framed as a replacement for
adaptive HPO systems. Hyperband, ASHA, BOHB, and Bayesian optimization
address broader optimizer problems. This paper addresses a narrower
question: can a small, transparent promotion schedule stop additional
long-horizon spending in a fixed experimental branch after declared
criteria are not met? In this case, yes.

Among the six candidates not promoted to \texttt{60} minutes, all had
higher \texttt{val\_bpb} than d8/ar48 at the replicated
\texttt{10}-minute gate, and d8/ar48 itself did not meet the
cheaper-architecture threshold at \texttt{12} hours. This transitivity
argument does not prove the dropped conditions would have failed, but it
bounds the plausibility gap under the observed gate sequence.

\section{7. Limitations}\label{7-limitations}

The long-horizon confirmation has only two seeds per final condition.
The result is a descriptive blocked comparison, not a high-powered
inferential study.

Seeds and budgets are partially confounded across stages. The
\texttt{10}-minute, \texttt{60}-minute, and \texttt{12}-hour reads use
different seed ranges, so ranking changes across stages cannot be
attributed to training duration alone. The stage comparisons are
operational promotion evidence, not within-seed learning-curve
estimates.

The hosts are heterogeneous. Windows A100 and Linux L40S should be read
as host blocks, not matched hardware replications.

The \texttt{12}-hour comparison is also capacity-confounded. The bridge
reference is the largest of the three final conditions, and the paper
does not include equal-parameter or equal-token controls that would
separate model capacity from promotion-rule effects.

The run order was sequential. Thermal, cache, filesystem, or host-state
effects could influence timing and throughput. The main endpoint is
final \texttt{val\_bpb}, but order should still be disclosed in any
public artifact package.

The budget counterfactuals are accounting comparisons. They do not prove
that uncontinued candidates would fail at \texttt{12} hours.

The candidate matrix is inherited from a prior branch and local variants
around it. The result does not establish global optimality for the
configuration space.

The paper does not run Hyperband, ASHA, BOHB, or Bayesian optimization
baselines. Those methods are related work and framing context, not
defeated comparators.

The public package uses curated summaries and source snapshots rather
than host-local execution folders. Scrubbed copies of the copied host
summaries are included under \texttt{anc/remote-results/}, with
path-bearing columns removed. Internal \texttt{remote-plans/}, raw run
directories, checkpoint paths, and launcher scripts contain
environment-specific paths and are excluded from the arXiv upload
bundle.

\section{8. Conclusion}\label{8-conclusion}

In this fixed micro-pretraining runner and twelve-condition candidate
matrix, a conservative staged promotion rule retained the observed
long-horizon bridge reference in the promoted set across two
\texttt{12}-hour seeds and two heterogeneous host blocks; that final
comparison remains capacity-confounded because the bridge reference is
also the largest final condition. The same frozen thresholds were not
met by a greedy comparator or a plausible cheaper-architecture branch,
so the study stopped before spending an additional \texttt{24}-hour or
continue-all budget.

The paper\textquotesingle s main lesson is disciplined stopping. Small
experiments add value when they are used to decide what not to run next,
provided the promotion rule retains reference conditions, repeats
unstable cheap screens, and avoids turning short-budget leaders into
unsupported long-horizon claims. The portable part is the discipline:
predeclared reference/control roles, replicated cheap screens, frozen
stopping rules, explicit stopping, and transparent sensitivity checks.
The runner-specific parts are the absolute \texttt{0.010} and
\texttt{0.020} thresholds, the all-cell versus mean-gap threshold
asymmetry, the A100/L40S host block, and the \texttt{169.2} GPU-hour
accounting.

\section{Appendix A. Reproducibility
Snapshot}\label{appendix-a-reproducibility-snapshot}

The public arXiv package uses a curated \texttt{anc/} directory rather
than host-local execution folders. Start with:

\begin{itemize}
\tightlist
\item
  \texttt{anc/README.txt} for the package overview.
\item
  \texttt{anc/MANIFEST.json} for the package inventory.
\item
  \texttt{anc/table\_manifest.json} to map manuscript tables to data.
\item
  \texttt{anc/figure\_manifest.json} to map figures to source artifacts.
\end{itemize}

The internal project workspace is organized around four artifact groups:

\begin{itemize}
\tightlist
\item
  Candidate definition and gates: the starting matrix plus frozen
  pre-analysis and stage-decision records.
\item
  Observed run summaries: copied host summaries and derived
  stage-analysis tables, with path-bearing columns removed from the
  public bundle.
\item
  Reproducibility scripts: the figure generator, analysis scripts, and
  instrumented runner snapshot.
\item
  Public figures and manifests: manuscript figures plus machine-readable
  mappings from figures and tables back to their source artifacts.
\end{itemize}

The public package includes source snapshots, scrubbed matrices,
scrubbed remote-result summaries, scrubbed analysis summaries,
figure-generation scripts, and separate figure/table manifests that map
derived public outputs to source artifacts or declared setup records.
The gate records are under \texttt{anc/preanalysis/}; threshold
sensitivity is under
\texttt{anc/analysis/p06\_threshold\_sensitivity\_2026-05-02.*}; and
host-result summaries are under \texttt{anc/remote-results/}.

See Section 4.1 for the fixed runner, dataset, validation shard,
tokenizer, context length, and validation-token count.

The runner source snapshot records the seed calls and CUDA environment
fields captured by the summary writer. Deterministic PyTorch algorithms
are not enabled in the public runner snapshot, so exact bitwise replay
across host blocks is not claimed.

The generated cache binaries are not bundled in this paper workspace.
The public source bundle provides the source snapshot, generator code,
dataset/shard identifiers, tokenizer artifact names, scrubbed host
summaries, and derived analysis tables. Exact binary cache hashes are
therefore unavailable unless recovered from the training hosts.

The local internal folder contains remote execution plans and copied
summaries with environment-specific paths. These are useful for
provenance, but the public arXiv package uses curated summaries and
excludes those host-local operational files. The bundle also excludes
Python caches, raw run work directories, host-local launch scripts,
checkpoint binaries, and absolute private host paths. The included
scrubbed summaries are sufficient to re-derive the reported tables and
figures without the private execution folders.

\section{References}\label{references}

{[}1{]} Felipe Chavarro Polania. Staged Factorial Screening for
Budget-Constrained Micro-Pretraining. arXiv:2606.05186, 2026.
\url{https://arxiv.org/abs/2606.05186}

{[}2{]} Lisha Li, Kevin Jamieson, Giulia DeSalvo, Afshin Rostamizadeh,
and Ameet Talwalkar. Hyperband: A Novel Bandit-Based Approach to
Hyperparameter Optimization. Journal of Machine Learning Research,
18(185):1-52, 2018. \url{https://jmlr.org/papers/v18/16-558.html}

{[}3{]} Liam Li, Kevin Jamieson, Afshin Rostamizadeh, Ekaterina Gonina,
Jonathan Ben-tzur, Moritz Hardt, Benjamin Recht, and Ameet Talwalkar. A
System for Massively Parallel Hyperparameter Tuning. Proceedings of
Machine Learning and Systems, 2020.
\url{https://proceedings.mlsys.org/paper_files/paper/2020/hash/a06f20b349c6cf09a6b171c71b88bbfc-Abstract.html}

{[}4{]} Stefan Falkner, Aaron Klein, and Frank Hutter. BOHB: Robust and
Efficient Hyperparameter Optimization at Scale. Proceedings of the 35th
International Conference on Machine Learning, PMLR 80:1437-1446, 2018.
\url{https://proceedings.mlr.press/v80/falkner18a.html}

{[}5{]} Jesse Dodge, Suchin Gururangan, Dallas Card, Roy Schwartz, and
Noah A. Smith. Show Your Work: Improved Reporting of Experimental
Results. Proceedings of EMNLP-IJCNLP, pages 2185-2194, 2019.
\url{https://aclanthology.org/D19-1224/}

{[}6{]} Ian Magnusson, Nguyen Tai, Ben Bogin, David Heineman, Jena D.
Hwang, Luca Soldaini, Akshita Bhagia, Jiacheng Liu, Dirk Groeneveld,
Oyvind Tafjord, Noah A. Smith, Pang Wei Koh, and Jesse Dodge.
DataDecide: How to Predict Best Pretraining Data with Small Experiments.
arXiv:2504.11393, 2025. \url{https://arxiv.org/abs/2504.11393}

{[}7{]} Kaiyue Wen, David Hall, Tengyu Ma, and Percy Liang. Fantastic
Pretraining Optimizers and Where to Find Them. arXiv:2509.02046, 2025.
\url{https://arxiv.org/abs/2509.02046}

\end{document}